\documentclass[pmlr]{jmlr}% new name PMLR (Proceedings of Machine Learning)

\usepackage{longtable}% for long tables
\usepackage{multirow}

\usepackage{booktabs}
\usepackage[load-configurations=version-1]{siunitx} % newer version
\usepackage{float}
\usepackage{pifont}

\makeatletter
\def\set@curr@file#1{\def\@curr@file{#1}} %temp workaround for 2019 latex release
\makeatother

\theorembodyfont{\upshape}
\theoremheaderfont{\scshape}
\theorempostheader{:}
\theoremsep{\newline}

\usepackage{enumitem}

\title[COVID-CT-Dataset: A CT Image Dataset about COVID-19]{COVID-CT-Dataset: A CT Image Dataset about COVID-19}

\author{\Name{Xingyi Yang}
       \Email{x3yang@eng.ucsd.edu}\\
       \addr UC San Diego
\AND
\Name{Xuehai He}
       \Email{x5he@eng.ucsd.edu}\\
       \addr UC San Diego
\AND
\Name{Jinyu Zhao}
       \Email{jiz077@eng.ucsd.edu} \\
       \addr UC San Diego
       \AND
 \Name{Yichen Zhang}
       \Email{yiz037@eng.ucsd.edu}\\
       \addr UC San Diego
       \AND
       \Name{Shanghang Zhang}
       \Email{shz@eecs.berkeley.edu}\\
       \addr UC Berkeley
       \AND
       \Name{Pengtao Xie}
       \Email{pengtaoxie2008@gmail.com}\\
       \addr UC San Diego
       } 

\begin{document}

\maketitle

\begin{abstract}
	During the outbreak time of COVID-19, computed tomography (CT) is a useful manner for diagnosing COVID-19 patients. Due to privacy issues, publicly available COVID-19 CT datasets are highly difficult to obtain, which hinders the research and development of AI-powered diagnosis methods of COVID-19 based on CTs. To address this issue, we build an open-sourced dataset – COVID-CT, which contains 349 COVID-19 CT images from 216 patients and 463 non-COVID-19 CTs. The utility of this dataset is confirmed by a senior radiologist who has been diagnosing and treating COVID-19 patients since the outbreak of this pandemic. We also perform experimental studies which further demonstrate that this dataset is useful for developing AI-based diagnosis models of COVID-19. Using this dataset, we develop diagnosis methods based on multi-task learning and self-supervised learning, that achieve an F1 of 0.90, an AUC of 0.98, and an accuracy of 0.89.  According to the senior radiologist, models with such performance are good enough for clinical usage. The data and code are available at \url{https://github.com/UCSD-AI4H/COVID-CT}
 % CT scans are promising in providing accurate, fast, and cheap screening and testing of COVID-19. In this paper, we build a publicly available COVID-CT dataset, containing 349 CT scans that are positive for COVID-19, to foster the research and development of deep learning methods which predict whether a person is affected with COVID-19 by analyzing his/her CTs. We train several deep convolutional neural network on this dataset and the best network achieves an F1 of 0.85 which is a promising performance but yet to be further improved.
\end{abstract}

\section{Introduction} 

%\footnotetext{*Equal contribution}

Coronavirus disease 2019 (COVID-19) is an infectious disease that has caused about 386,000 deaths all over the world, among 6.5 million infected cases, as of June 3rd, 2020. 
%more than $2.9$ million individuals all over the world and caused more than $200,000$ deaths, as of April 26 in 2020. 
One major hurdle in controlling the spreading of this disease is the shortage of tests. The current tests are mostly based on reverse transcription polymerase chain reaction (RT-PCR). During the peak time of COVID-19 outbreak, RT-PCR test kits were in great shortage. As a result, many suspected cases cannot be tested in time and they continue to spread the disease to others  unconsciously.
%It takes 4-6 hours to obtain results, which is a long time compared with the rapid spreading rate of COVID-19. Besides inefficiency, RT-PCR test kits are in huge shortage. 
To mitigate the shortage of RT-PCR test kits, hospitals have been utilizing alternative diagnosis methods. Among them, computed tomography (CT) scans have been used for screening and diagnosing COVID-19. For example, in the Diagnosis and Treatment Protocol for Novel
Coronavirus Pneumonia (Trial Version 5)\footnote{\url{http://www.kankyokansen.org/uploads/uploads/files/jsipc/protocol_V5.pdf}} made by the National Health Commission and  State Administration of Traditional Chinese Medicine in China, CT and other chest imaging techniques have been named as an important way for diagnosing COVID-19. Following this guideline, many hospitals in China used CT scans for COVID-19 diagnosis, which has been demonstrated to be effective. 

To further understand the role of CTs in diagnosing COVID-19, we consulted a senior radiologist in Tongji Hospital, Wuhan, China, who has been diagnosing and treating COVID-19 patients since the outbreak of this disease in China. According to this radiologist, during the outbreak time, CT is useful for diagnosing COVID-19; out of the outbreak time, CT is not as useful. The reasons are as follows. CTs can be used to judge whether a patient is infected by viral pneumonia (COVID-19 is a type of viral pneumonia caused by the  SARS-CoV-2 virus). However, CTs do not have the ability to determine which virus is causing the viral pneumonia: SARS-CoV-2 or others. Strictly speaking, CTs cannot be used to confirm whether a patient is infected by COVID-19. However, during the outbreak time, most viral pneumonia is caused by SARS-CoV-2. That is to say, if a patient is confirmed to have viral pneumonia according to CT results, this viral pneumonia is very likely to be COVID-19. Due to this fact, CTs are considered useful for diagnosing COVID-19 during its outbreak.

During the outbreak of COVID-19, radiologists are highly occupied, who may not have the bandwidth to read a number of CT scans timely. Besides, for radiologists in under-developed areas, such as rural regions, they may not be well-trained to recognize COVID-19 from CT scans, since this disease is relatively new. To address these problems, several works~\citep{huang2020serial,li2020artificial} have developed deep learning methods to screen COVID-19 from CTs. 
%To address this problem, several works have developed deep learning methods 
%There have been several works developing deep learning methods to analyze CT scans for 
Due to privacy concerns, the CT scans used in these works are not shared with the public. This greatly hinders the research and development of more advanced AI methods for more accurate screening of COVID-19 based on CTs.

To address this issue, we built a COVID-CT  dataset which contains 349 CT images positive for COVID-19 belonging to 216 patients and 397 CT images that are negative for COVID-19. The dataset is open-sourced to the public, to foster the R\&D of CT-based testing of COVID-19. From 760 medRxiv and bioRxiv preprints about COVID-19, we extract reported CT images and manually select those containing clinical findings of COVID-19 by reading the captions of these images. After releasing this dataset, we received several feedback expressing concerns about the usability of this dataset. The major concerns are summarized as follows. First, when the original CT images are put into papers, the quality of these images is degraded, which may render the diagnosis decisions less accurate. The quality degradation includes: the Hounsfield unit (HU) values are lost; the number of bits per pixel is reduced; the resolution of images is reduced. Second, the original CT scan contains a sequence of CT slices, but when put into papers, only a few key slices are selected, which may have a  negative impact on diagnosis as well. 

We consulted the aforementioned radiologist at Tongji Hospital regarding these two concerns. According to the radiologist, the issues raised in these concerns do not significantly affect the accuracy of diagnosis decision-making. First,  experienced radiologists are able to make accurate diagnosis from low quality CT images. For example, given a photo taken by smartphone of the original CT, experienced radiologists can make accurate diagnosis by just looking at the photo, though the CT image in the photo has much lower quality than the original CT. Likewise, the quality gap between CT images in papers and original CTs will not largely hurt the accuracy of diagnosis. Second, while it is preferable to read a sequence of CT slices, oftentimes a single-slice of CT contains enough clinical information for accurate decision-making. 

%single-slice CT images contain usefu

%Our consulting radiologist from Tongji Hospital told us experienced radiologists can make diagnosis and treatment decisions based on CT images extracted from papers and such decisions are as accurate as those made based on original CT scans. The reduced image quality and reduced number of slices won't largely affect the correctness of clinical decisions.

%After releasing this dataset, we received constructive feedback expressing concerns that there might be a large domain difference between CT images extracted from papers and original CT scans. One major concern is that CT images extracted from papers have lower quality (e.g., lower resolution, lower bits, loss of HU values) than the original images. Another concern is CT images in papers are selected single-slice CTs whereas original CT scans are a sequence of CT slices. 

To further address these concerns, we use the CT images extracted from papers  only  for model training, not for evaluation.  For testing, we use original CT images donated by hospitals.  Validation is also conducted on original CTs. We compare models trained on our paper-extracted CTs and models trained on original CTs, and the former outperforms the latter. This demonstrates that COVID-19 CTs extracted from papers are useful for training diagnosis models of COVID-19. In the end, we leverage our COVID-CT dataset, original CT images collected elsewhere, lesion masks labeled by radiologists to train a diagnosis model of COVID-19, based on multi-task learning and self-supervised learning. This model achieves an F1 of 0.90, an AUC of 0.98, and an accuracy of 0.89. According to the senior radiologist, models with such performance are good enough for clinical usage.   

The major contributions of this paper are as follows: (1) We collect a COVID-19 CT dataset, which contains 349 positive COVID-19 CT images from 216 patients; (2) We verify the usefulness of this dataset for  developing COVID-19 diagnosis models via experiment studies; (3) We develop methods based on multi-task learning and contrastive self-supervised learning to improve the diagnosis accuracy to a clinically useful level.

The rest of the paper is organized as follows. In Section 2, we introduce the COVID-CT dataset. In Section 3, we perform a study to verify whether COVID-CT is useful for training COVID-19 diagnosis models. In Section 4, we develop methods based on multi-task learning and self-supervised learning to improve the diagnosis accuracy of COVID-19. Section 5 reviews related works and Section 6 concludes the paper.

%We trained several deep learning models on 191 COVID CTs  and 234 Non-COVID CTs to predict whether a CT image is positive for COVID-19.  Tested on 98 COVID CTs and 105 Non-COVID CTs, our best model achieves an F1 score of 0.85. The results demonstrate that CT scans are promising for screening and testing COVID-19, while more advanced methods are needed to further improve the accuracy. 

\section{The COVID-CT Dataset}

\begin{figure}[h]
	\begin{center}
 	\includegraphics[width = 0.4\columnwidth]{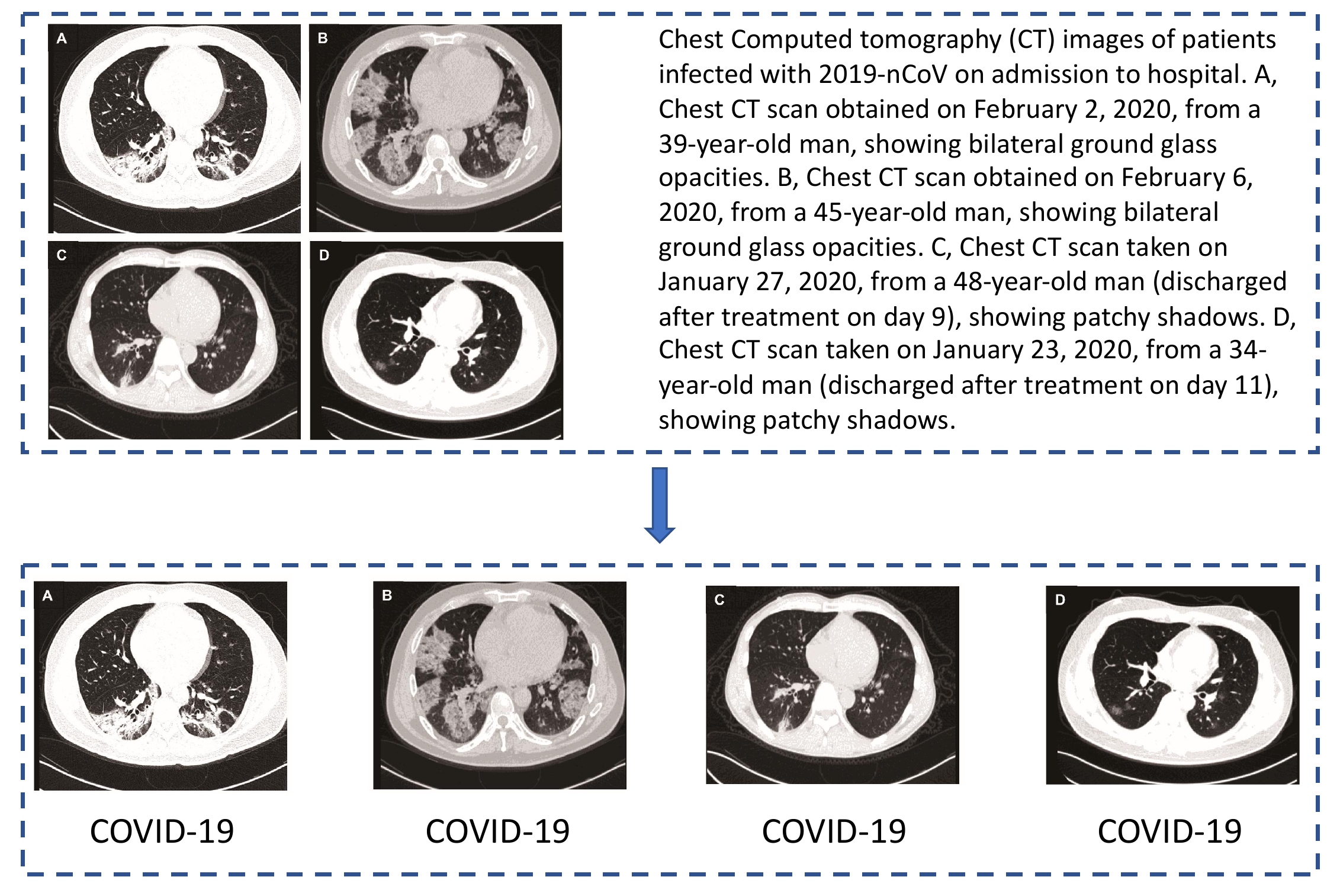}
 	\hspace{0.1cm}
 	\includegraphics[width=0.51\textwidth]{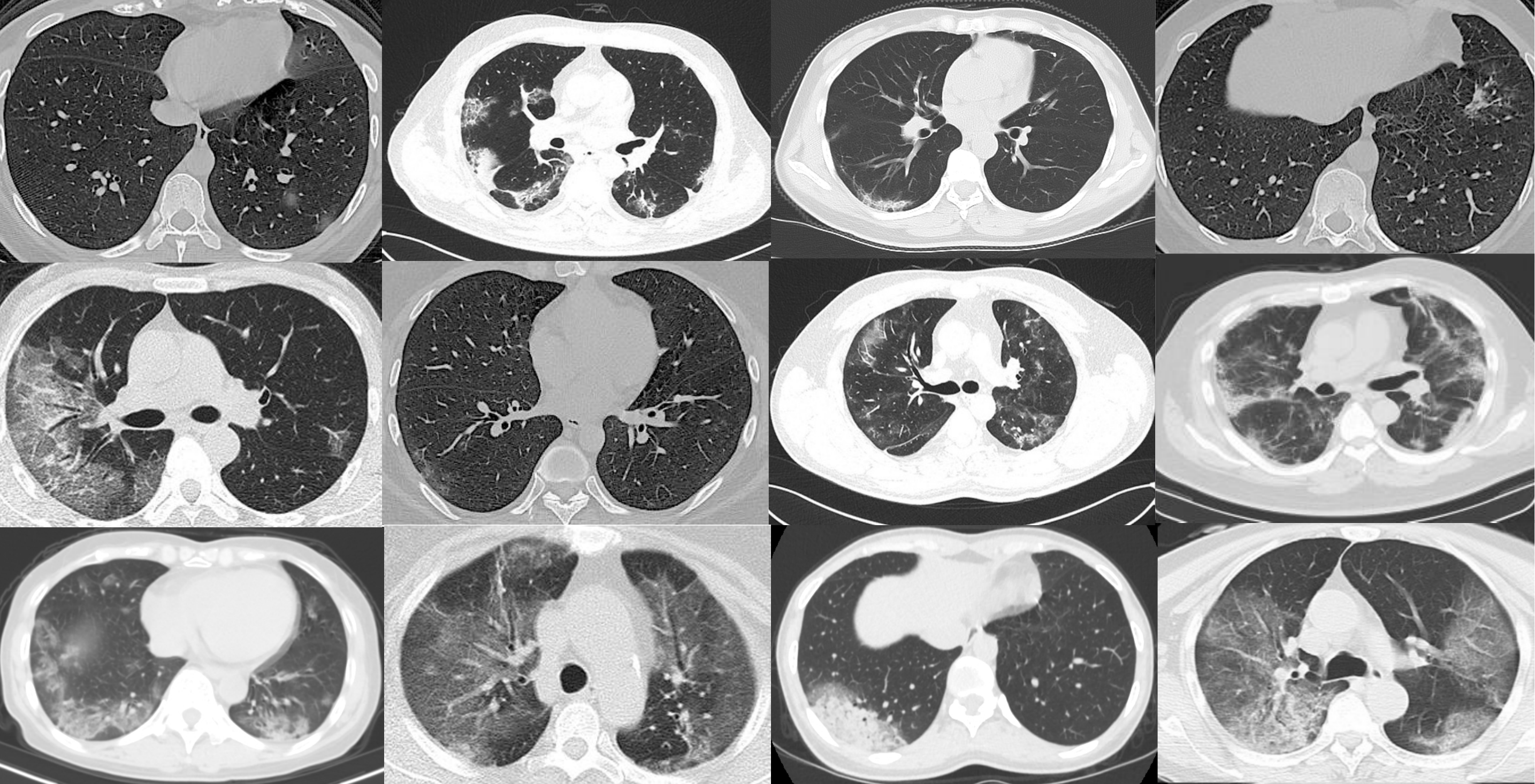}
 	\caption{(Left) For any figure that contains multiple CT images as sub-figures, we manually split it into individual CTs. (Right)  Examples of CT images that are positive for COVID-19.
	}\label{fig:ct-subfig}
	\end{center}
 \end{figure}

In this section, we describe how the COVID-CT dataset is built. We first collected 760 preprints about COVID-19 from medRxiv\footnote{https://www.medrxiv.org/} and bioRxiv\footnote{https://www.biorxiv.org/}, posted from Jan 19th to Mar 25th. Many of these preprints report patient cases of COVID-19 and some of them show CT images in the reports. CT images are associated with  captions describing the clinical findings in the CTs. We  used PyMuPDF\footnote{https://github.com/pymupdf/PyMuPDF} to extract the low-level structure information of the PDF files of preprints and located all the embedded figures. The quality (including resolution, size, etc.) of figures were well-preserved. From the structure information, we also identified the captions associated with figures. Given these extracted figures and captions, we first manually selected all CT images. Then for each CT image, we read the associated caption to judge whether it is positive for COVID-19. If not able to judge from the caption, we located the text analyzing this figure in the preprint to make a decision. For each CT image, we also collected the meta information extracted from the paper, such as patient age, gender, location, medical history, scan time, severity of COVID-19, and radiology report. For any figure that contains multiple CT images as sub-figures, we manually split it into individual CTs, as shown in Figure~\ref{fig:ct-subfig}(left).

\begin{figure}[h]
	\begin{center}
 	\includegraphics[width = 0.55\columnwidth]{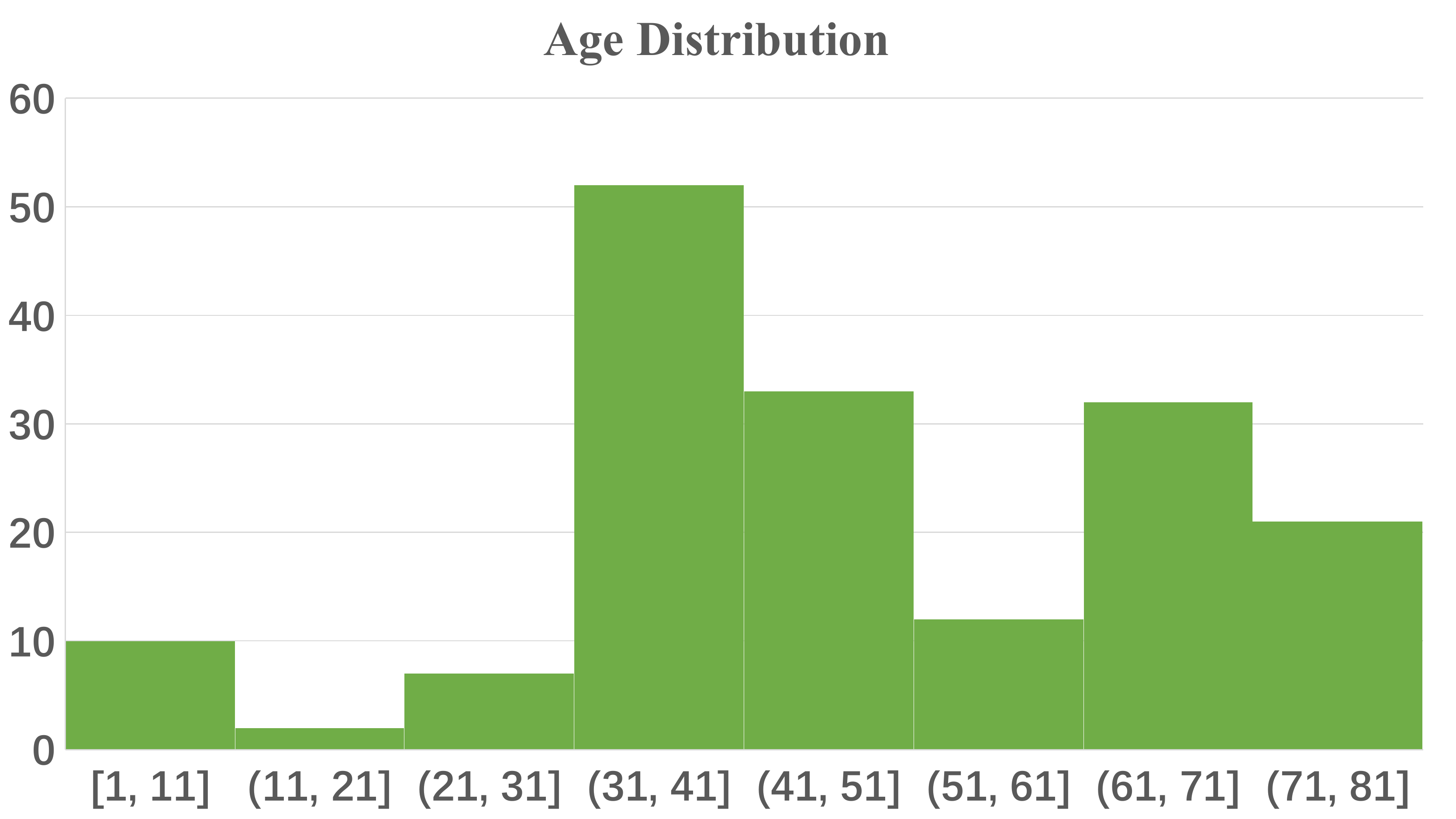}
 	\hspace{0.2cm}
 	\includegraphics[width=0.25\columnwidth]{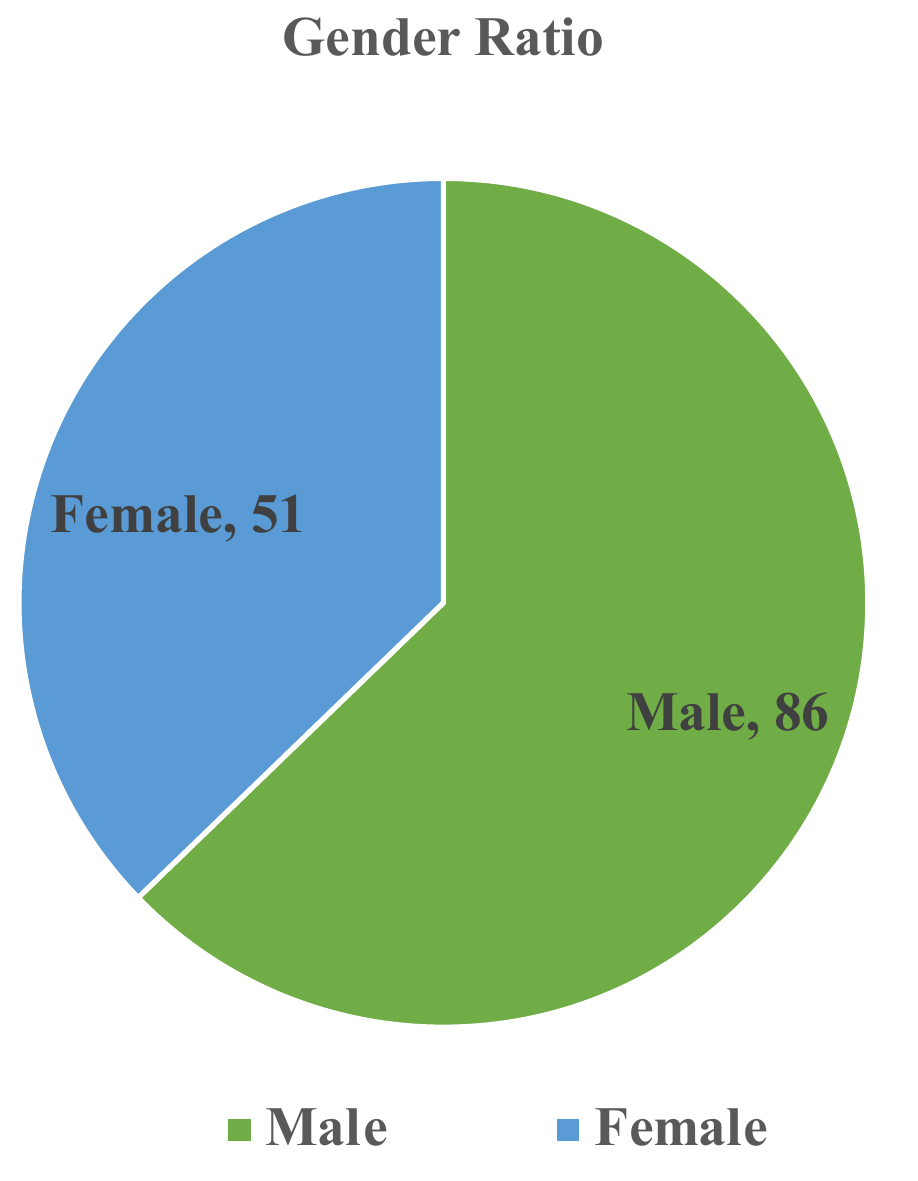}
 	\caption{(Left) Age distribution of COVID-19 patients.
 	(Right) The gender ratio of  COVID-19 patients. The ratio of male:female is 86:51.
	}\label{fig:age-stats}
	\end{center}
 \end{figure}

In the end, we obtain 349 CT images labeled as being positive for COVID-19. These CT images have different sizes. The minimum, average, and maximum height are 153, 491, and 1853. The minimum, average, and maximum width are 124, 383, and 1485. These images are from 216 patient cases. 
Figure~\ref{fig:ct-subfig}(right) shows some examples of the COVID-19 CT images. For patients labeled with positive, 169 of them have age information and 137 of them have gender information.
Figure~\ref{fig:age-stats}(left) shows the age distribution of patients with COVID-19. 
Figure~\ref{fig:age-stats}(right) shows the gender ratio  of patients with COVID-19. Male patients are more than female patients, with a number of 86 and 51 respectively. 
Table~\ref{tab:compare} compares the COVID-CT dataset with others. Our dataset has more COVID19-positive images and patients than COVID-19 Image Data Collection and SIRM COVID-19 Database. COVID-CS is not open-sourced. COVID-19 CT Segmentation Dataset has more COVID19-positive images, but less patients. Note that CT images of the same patient are highly similar visually. Therefore, the diversity of images in our dataset is larger than that in the  COVID-19 CT Segmentation Dataset.

\begin{table}[t]
    \centering
    \small
    \caption{Comparison of COVID-CT with other datasets.}
\begin{tabular}{p{5cm}ccccc}
\toprule Dataset   & \# COVID images & \# patients & Open-source& Metadata \\
\midrule \textbf{COVID-CT}  & 349 & 216 & Y & Y \\
\hline
\shortstack[l]{COVID-19 Image Data Collection \\\citep{c1}}  & 84 & 45 & Y & Y \\
\hline
 SIRM COVID-19 Database~\footnote{https://www.sirm.org/
category/senza-categoria/covid-19/} & 100 & 60 & Y &  Y \\
\hline
 COVID-CS~\citep{wu2020jcs} & 68626   &   400 &   N &N\\ 
 \hline
 COVID-19 CT Segmentation Dataset~\citep{jun1covid} & 379 & 20  & Y &  N \\
\bottomrule
\end{tabular}
\label{tab:compare}
\end{table}

\iffalse
 \begin{figure}[H]
	\begin{center}
 	\includegraphics[width = 0.5\columnwidth]{figs/COVID_CT.pdf}
 	\caption{Examples of CT images that are positive for COVID-19.
	}\label{fig:examples}
	\end{center}
 \end{figure}
\fi

\paragraph{Collection of non-COVID-19 CT images as negative training set} To develop binary classification models for diagnosing COVID-19, in addition to the 349 COVID-19 CT images, we also collect a set of non-COVID-19 CT images as negative training examples.  The sources of these images include:
\begin{itemize}[leftmargin=*]
    \item The MedPix\footnote{https://medpix.nlm.nih.gov/home} database, which is an open-access online database of medical images, teaching cases, and clinical topics. It contains more than 9000 topics, 59000 images from 12000 patient cases.
    \item The LUNA\footnote{https://luna16.grand-challenge.org/} dataset, which contains 888 lung cancer CT scans from 888 patients.
    \item  The Radiopaedia website\footnote{https://radiopaedia.org/articles/covid-19-3}, which contains radiology images from 36559 patient cases.
        \item PubMed Central (PMC)\footnote{https://www.ncbi.nlm.nih.gov/pmc/}, which is a free full-text archive of biomedical and life sciences journal literature. Some papers contain CT images. 
\end{itemize}

\iffalse
\begin{figure}[H]
\centering
\begin{minipage}[H]{0.45\textwidth}
\centering
\includegraphics[width=7.5cm]{figs/age.pdf}
\caption{Age distribution of COVID-19 patients}
\label{fig:age}
\end{minipage}
\hspace{0.2cm}
\begin{minipage}[H]{0.45\textwidth}
\centering
\includegraphics[width=4.8cm]{figs/gender.pdf}
\caption{ }
\label{fig:gender}
\end{minipage}
\end{figure}
\fi

\begin{table}[h]
    \centering
    % \footnotesize
\caption{Statistics of the negative training set}
\begin{tabular}{ccccccc}
\toprule
&Class   &LUNA& MedPix&  PMC & Radiopaedia & Total \\
\midrule
\# patients & Non-COVID  &17 &55&55 &2&  55 \\
\# images&Non-COVID &36 &195 & 202 &30&  463 \\
\bottomrule
\end{tabular}
\label{tb:negtrain data}
\end{table}

Table~\ref{tb:test data} shows the composition of the negative training set which contains 463 images from 55 patients: 36 images from LUNA, 195 from MedPix, 202 from PMC, and 30 from Radiopaedia.

\paragraph{Collection of test and validation images}

To evaluate the trained models, we collect a validation set and a test set. In these two sets, all images are original CTs donated by hospitals. None of them is extracted from papers. These original COVID-19 CTs come from the COVID-19 CT Segmentation dataset (COVID-Seg)\footnote{http://medicalsegmentation.com/covid19/}, which contains 20 axial volumetric COVID-19 CT scans. The sources of original non-COVID-19 CTs include LUNA and Radiopaedia. Table~\ref{tb:test data} shows the composition of the test set. It has 173 COVID-19 CT images, from 4 patients in the COVID-Seg dataset. It contains 168 non-COVID-19 CT images: 164 of them are from 19 patients in the LUNA dataset and the rest 4 are from 1 patient in Radiopaedia. 

\begin{table}[t]
    \centering
    \caption{Statistics of the test set}
    \begin{tabular}{cccccc}
    \toprule
    &Class   &LUNA& COVID-Seg&Radiopaedia & Total \\
    \midrule
    \multirow{2}{*}{\# patients}&COVID  &0 &4& 0 & 4 \\
    % \cmidrule(r){2-6}
    &Non-COVID & 19   &0&1 & 20\\
    \midrule
    \multirow{2}{*}{\# images}&COVID &0  &173& 0 & 173\\
    % \cmidrule(r){2-6}
    &Non-COVID & 164   &0&4 & 168\\
    \bottomrule
    \end{tabular}
    \label{tb:test data}
\end{table}

\begin{table}[t]
     \centering
    % \footnotesize
    \caption{Statistics of the validation set}
    \begin{tabular}{cccccc}
    \toprule
    &Class   &LUNA & COVID-Seg&Radiopaedia & Total\\
    \midrule
    \multirow{2}{*}{\# patients}&COVID  &0 &4& 0 & 4\\
    % \cline{2-6}
    &Non-COVID & 38   &0&1 & 39\\
    \midrule
    \multirow{2}{*}{\# images}&COVID  &0 &88& 0 & 88\\
    % \cline{2-6}
    &Non-COVID & 48   &0&16 & 64\\
    \bottomrule
    \end{tabular}
    \label{tb:val data}
\end{table}
% \begin{table}[t]
% \end{table}
% \begin{table}[t]
%     \centering
%     % \footnotesize
% \caption{Statistics of the validation set}
% \scalebox{0.85}{
% \begin{tabular}{cccccc}
% \toprule
% &Class   &LUNA & COVID-Seg&Radiopaedia & Total\\
% \midrule
% \multirow{2}{*}{\# patients}&COVID  &0 &4& 0 & 4\\
% % \cline{2-6}
% &Non-COVID & 38   &0&1 & 39\\
% \midrule
% \multirow{2}{*}{\# images}&COVID  &0 &88& 0 & 88\\
% % \cline{2-6}
% &Non-COVID & 48   &0&16 & 64\\
% \bottomrule
% \end{tabular}}
% \label{tb:val data}
% \end{table}
Table~\ref{tb:val data} shows the composition of the validation set. It has 88 COVID-19 CT images, from 4 patients in the COVID-Seg dataset. It contains 64 non-COVID-19 CT images: 48 of them are from 38 patients in the LUNA dataset and the rest 16 are from 1 patient in Radiopaedia. The  images from LUNA are either about lung cancer or normal. The images from Radiopaedia are normal. We automatically crop the chest position in images through threshold segmentation and convex hull detection, which gets rid of white space at the edge of images.

\section{Study I: Is COVID-CT useful for training CT-based diagnosis models of COVID-19?}
As mentioned before, we received several feedback expressing concerns that the COVID-CT dataset may not be useful for training CT-based diagnosis models for COVID-19 since (1)  images in COVID-CT are extracted from papers in PDF format, which are likely to have lower quality compared with original CT scans and (2) images in COVID-CT are single-slice CTs rather than multiple-slice CTs as in original CT scans. In this section, we perform an experimental study to investigate whether the images in COVID-CT are useful for training CT-based diagnosis models.

\paragraph{Study Design} We compare experimental settings with three different positive training datasets. 
\begin{itemize}[leftmargin=*]
    \item COVID-Seg: We use 118 positive CTs in COVID-Seg as positive training examples. These images are original CT images obtained from the image archiving systems in hospitals.
    \item COVID-CT-349: We use the full set of 349 COVID-19 CTs in COVID-CT as positive training examples. These images are extracted from papers. 
    \item COVID-CT-118: We randomly sample 118 images from  COVID-CT as positive training examples. These images are extracted from papers.
  %  \item Combined: We combine the 349 images in COVID-CT-349 and the 118 images in CT-Seg. The combined dataset contains 467 images. 
\end{itemize}

The negative training images, test images, and  validation images are the same for the above-mentioned three settings, as given in Table~\ref{tb:negtrain data}, \ref{tb:test data}, and \ref{tb:val data} respectively.  
%are the same across three settings. 
%They are original CT images obtained from the image archiving systems in hospitals. 
The images used for validation and testing are original CT images obtained from the image archiving systems in hospitals. 
%\subsection{Methods} 
%We develop three methods for CT-based diagnosis of COVID-19. 
%\begin{itemize}
%    \item DenseNet-169: The network is pretrained on the ImageNet dataset.
%    \item 
%\end{itemize}
We use the DenseNet-169~\citep{Huang_2017_CVPR} and ResNet-50~\citep{he2016deep} models for classifying a CT image as COVID-19 or non-COVID-19. They are pretrained on the ImageNet~\citep{imagenet} dataset. The trained models are evaluated  using three metrics: accuracy, F1, and area under ROC curve (AUC). For all metrics, the higher the better. Hyperparameters are tuned on the validation set. 

\paragraph{Experimental Settings}
Input images are resized to 480-by-480. We perform data augmentation on the training set. Each training image is augmented with random cropping with a scale of 0.5, horizontal flip, random contrast, and random brightness with a factor of 0.2. The weight parameters in the networks are optimized using Adam~\citep{adam} with an initial learning rate of 0.0001 and a mini-batch size of 16. Cosine scheduling with a period of  10 is used to  adjust  the  learning  rate  across  the  training  process. We implement the network using PyTorch and train it on one GTX 1080Ti GPU. 

%We evaluate our method using three metrics: (1) Accuracy; (2) F1-score~\citep{F1}; (3) Area under ROC curve (AUC)~\citep{auc}. For all metrics, the higher the better.

%\subsection{Results}
\paragraph{Results} Table~\ref{tab:Baseline} shows the results on the test set, under three settings of positive training images. 
%From Table~\ref{}, we make the following observations. First, COVID-CT-349 outperforms COVID-Seg substantially. This demonstrates that CT images extracted from papers are useful in training CT-based COVID-19 diagnosis models, given the fact that images in COVID-CT-349 are all extracted from papers while those in  COVID-Seg are original CT images. Second, COVID-Seg performs better than COVID-CT-118, where these two datasets have the same number of images. This indicates that original CTs are more useful that CTs extracted from papers. If one has access to a large number of original CTs, he or she should use original CTs for model training. However, this is not the case in the COVID-19 situation. 
From this table, we observe the following. First, the model trained on  COVID-CT-349 is largely better than that trained on COVID-Seg. COVID-CT-349 contains images extracted from papers. COVID-Seg contains original CT images. This demonstrates that COVID-CT is useful for training CT-based diagnosis models for COVID-19, despite the concerns that images extracted from papers have low quality and are single slices. Second, the model trained on COVID-CT-349 is much better than that trained on  COVID-CT-118. Adding more COVID-CT images into the training set substantially improves performance, which further demonstrates that the CTs in COVID-CT have high utility in training COVID-19 diagnosis models. Third, COVID-Seg outperforms COVID-CT-118. These two sets have the same number of images. This indicates that on average, an original  CT is more useful for model training than a CT extracted from papers. However, paper-extracted CTs are much easier to obtain than original CTs. In sum, these results demonstrate the usefulness of our COVID-CT dataset in training diagnosis models of COVID-19. 
%Since these images are publicly available, they can be used for free.
\begin{table}[htbp]
    \centering
        \caption{Performance of DenseNet-169 and ResNet-50 on the test set under different settings of positive training images.}
    \begin{tabular}{l|lccc}
    \toprule
   Model&  Positive training data & Accuracy (\%) & F1-score (\%) & AUC  (\%)
        \\ \midrule
        
  \multirow{3}{*}{DenseNet-169}  &   COVID-Seg & 69.8 & 61.1 & 86.9
        \\ 
  &      COVID-CT-349 & 79.5 &  76.0 & 90.1
       \\  
  &      COVID-CT-118 & 57.8 & 36.3 & 75.2\\
        \midrule
     %   \\ \hline 
%        Combined & 74.5 & 70.1
   \multirow{3}{*}{ResNet-50}     & COVID-Seg & 66.3 & 58.1 & 80.6
        \\ 
   &     COVID-CT-349 & 77.4 &  74.6 & 86.4
       \\  
    &    COVID-CT-118 & 60.4 &  42.6 & 74.1
        \\ \bottomrule 
    \end{tabular}
    \label{tab:Baseline}
\end{table}

%From Table~\ref{} and \ref{}, we observe similar results: (1) COVID-CT-349 outperforms COVID-Seg; (2) COVID-CT-349 outperforms COVID-CT-118; (3) COVID-CT-118 outperforms COVID-Seg. These results further demonstrate the usefulness of COVID-CT: CT images extracted from papers are useful in model training. 
%Comparing three tables, we make the following observations on methods. 

\section{Study II: Improve the performance to a clinically more useful level}
In the previous study, we have shown that paper-extracted COVID-CT is useful for training diagnosis models for COVID-19 and yields much better performance than purely using original COVID-19 CTs. But the accuracy is still low. In this section, we develop methods that improve the diagnosis accuracy to a clinically more useful level.  

%lower than 90\%, which is clinically not very useful. In this study, we investigate how to improve the performance to a level that meets the requirement in clinical practice. According to the radiologist from Tongji Hospital, an accuracy that is above 90\% is considered to be clinically useful. 

%\subsection{Method}
%Combining COVID-CT and 

%\subsection{Results}

\subsection{Method} 
\begin{figure}[h]
    \centering
     %\vspace{-0.3cm}
    \includegraphics[width=0.41\textwidth]{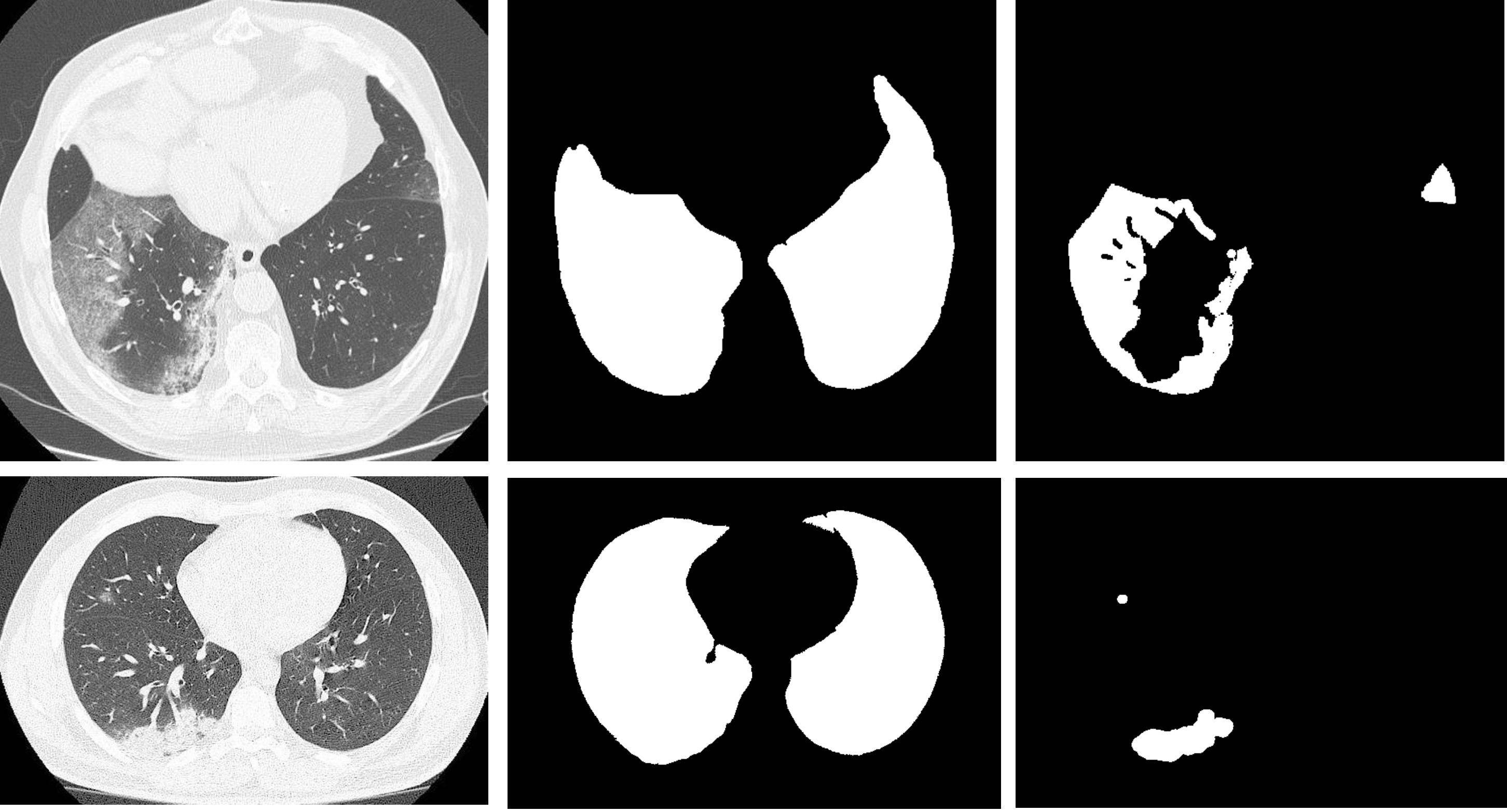}
    %\vspace{-0.3cm}
    \caption{CT images from COVID-Seg dataset (first column), lung masks (second column), and lesion masks (third column).}
    \label{fig:masks}
    %\vspace{-0.3cm}
\end{figure}

\begin{figure}[t]
    \centering
    \includegraphics[width=0.8\columnwidth]{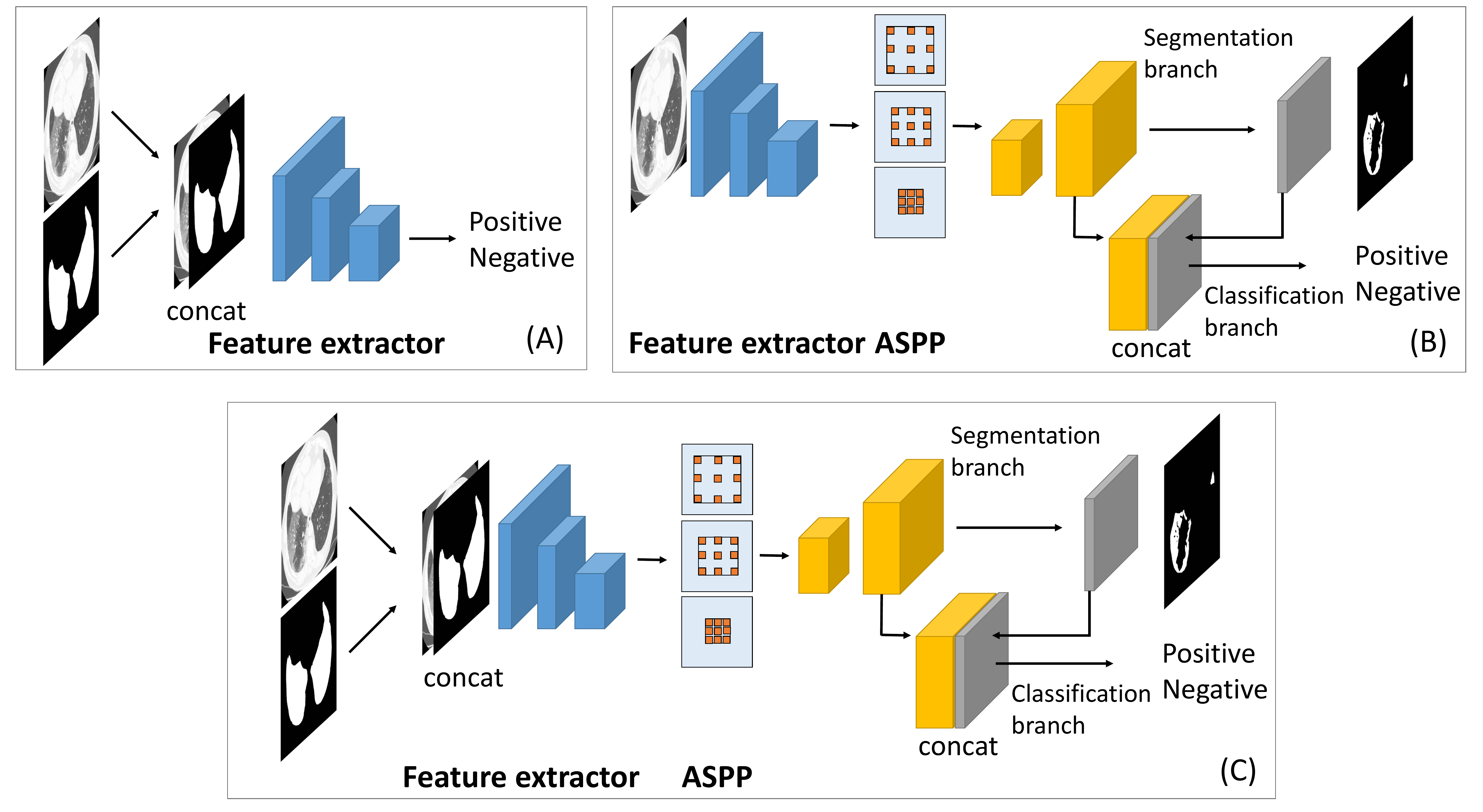}
    \caption{Architecture for our proposed models. (A) Architecture for incorporating lung masks as inputs. (B) Architecture for incorporating lesion masks as outputs. (C) Architecture for incorporating both lung masks and lesion masks. 
    %Illustration of the multitask network combining segmentation and classification branch. Both branch shares the same feature extractor.
    }
    \label{fig:Maskin_arch}
\end{figure}

The total number of positive training images in COVID-CT and COVID-Seg is 467. Training deep learning models on such a small number of images is prone to overfitting. To address this issue, we study two ways: one is to incorporate additional information including segmentation masks of lung regions and segmentation masks of lesion regions; the other way is to learn better visual representations.  Lung masks inform models to pay more attention to lung regions which contain clinical manifestation of COVID-19, instead of paying attention to background regions that are irrelevant to COVID-19. While lung masks narrow down the search space of COVID-19, they are still coarse-grained. Within the lung regions, areas exactly containing COVID-19 occupy a small proportion. Lesion masks can pinpoint such areas and provide fine-grained guidance to the model regarding which specific areas to pay attention to. While lesion masks are more accurate, they are more difficult to obtain. Only experienced radiologists can provide such labels while lung masks can be labeled by non-medical people. The positive and negative training images in COVID-CT and COVID-Seg are all labeled with lung masks. Only the positive training images in  COVID-Seg are labeled with lesion masks. Figure~\ref{fig:masks} shows some examples.  In light of these facts, we use different strategies to utilize lung masks and lesion masks. Lung masks are incorporated at the input side of the diagnosis model. Using the labeled lung masks in COVID-CT and COVID-Seg, we develop a lung segmentation model based on residual U-Net~\citep{zhang2018road}. Given a CT image (either during training or testing), we use the segmentation model to detect the lung mask, then concatenate it with the CT image and feed them into the diagnosis model (as shown in Figure~\ref{fig:Maskin_arch}(A)). The details of the lung segmentation experiments are deferred to the supplements. Lesion masks are incorporated at the output side of the diagnosis model during training (as shown in Figure~\ref{fig:Maskin_arch}(B)). They supervise the model to pay better attention to regions containing lesions. 

%Specifically, we incorporate the labeled masks of lung regions and lesion regions. Intuitively, the clinical manifestation of COVID-19 should be within the lung region. By incorporating lung masks, the model can pay more attention to the disease-containing lung regions instead of other non-lung regions. While the lung regions narrow down the search space, the majority of these regions are still disease-free. Masks of lesions relevant to COVID-19 provide more fine-grained and pinpointed guidance. 
Figure~\ref{fig:Maskin_arch}(C) shows the architecture for incorporating lung masks and lesion masks simultaneously. During training, given a CT image, the network  predicts the class label (whether this image is positive for COVID-19 or negative) and the lesion mask, by minimizing the sum of a classification loss and a lesion segmentation loss. Given a CT image, we first use the lung segmentation model to obtain the lung mask. Then the original CT image together with the lung mask image are concatenated and fed into the feature extraction network.  
%Figure~\ref{fig:MTN} shows the architecture of the proposed joint classification and segmentation framework. 
%Given the image, we first use DenseNet-169 to extract visual features. 
Then the visual features are fed into an atrous spatial pyramid pooling (ASPP)~\citep{chen2017rethinking} layer to extract denser and higher-resolution feature maps. These features maps are fed into two branches: a lesion segmentation branch for predicting lesion segmentation mask and a classification branch for prediction class label. The predicted lesion segmentation mask is fed into the classification branch as additional information. A lesion segmentation loss function is defined between the groundtruth mask and predicted mask. A classification loss is defined between the groundtruth label and predicted label. The sum of these two losses are minimized for learning the weight parameters in the network. Note that the lesion masks are only required during training. During testing, the prediction is made solely based on test images and their lung segmentation masks. No lesion mask is required.

The other approach for improving performance is to learn better visual representations. The weights in DenseNet-169 and ResNet-50 are pretrained on ImageNet using transfer learning. The class labels in ImageNet are leveraged during the pretraining. Since ImageNet classes are mostly non-medical, models pretrained by ImageNet may be biased to these general-domain classes and generalize less well on COVID-19 classification. To mitigate this bias, we use contrastive self-supervised learning (CSSL)~\citep{MoCo} to finetune the ImageNet-pretrained models on the training CT images in our COVID-19 diagnosis task. CSSL is an unsupervised approach. It learns representations by solving an auxiliary task. CSSL creates augmented examples of  CTs, then learns the visual representation network by solving the prediction task that whether two augments originate from the same  CT. During CSSL training, only the input CT images are used and no labels are utilized. After CSSL training, the network is further finetuned using both the input CTs and output labels.

%We incorporate lung mask as input to the predictive model. Figure~\ref{} illustrates the method. The original CT image and the lung mask are concatenated and fed into the prediction network as inputs. 

%These masks can inform the model to pay more attention to the lesion-containing regions. To incorporate these masks, we develop a multi-task learning framework that performs two tasks simultaneously: segment out the regions containing lesions and classify the images as COVID-19 or non-COVID-19. The masks are used to supervise the learning of the segmentation task. 

\subsection{Experiments}
\paragraph{Experimental settings}
%Following similar setting, all input images and masks are resized to $480\times 480$. Data augmentation like random cropping, horizontal flip, random rotation augmentation with angles uniformly drawn from $[-10^{\circ}10^{\circ}]$ is applied on both inpuut images and corresponding masks. The optimizer, learning rate and mini-batch size setup is consistent with experiments in section 3. It is worth noting that, for COVID-19 infection mask segmentation multi-task network, we only use the infection mask in the COVID-Seg dataset to train the segmentation branch. We implement the network using PyTorch and train it on one GTX 1080Ti GPU. 
In the multi-task learning framework, the weights associated with the segmentation loss and the classification loss are both set to 1. The rest of hyperparameter settings are the same as those in Section 3. In contrastive self-supervised learning, we follow the same hyperparameter setting as in MoCo~\citep{MoCo}. Each input image is augmented with random
horizontal flip, random cropping with a size of 0.2 in area,
random color jittering such as random brightness with a ratio
of 0.4, random contrast of 0.4, random saturation of 0.4,
random hue of 0.1, Gaussian blur, and random gray-scale
conversion. The size of the dynamic
dictionary was set to 512. Stochastic gradient descent (SGD)
was used as the optimizer with a minibatch size of 128, a weight decay of 0.0001, a momentum of 0.9, and an initial learning rate of 0.015.

%We compare the following five methods: (1) DenseNet-169 on COVID-Seg; (2) DenseNet-169 on COVID-CT-349; (3) DenseNet-169 on the combination of COVID-Seg and COVID-CT-349 (Combination), which contains x images; (4) Joint classification and segmentation (JCS) on COVID-Seg; (5) JCS on Combination.

%Additionally, lung semantic segmentation residual U-Net is trained with combination of the dice loss and cross-entropy loss to accurately locate the lung regions. The weights between two losses are selected to be 1:1. This model is trained on the input scale $320\times320$ with the mini-batch size of 16. We perform data augmentation on the training set. Each training image is augmented with random cropping with a scale between $[0.6,1.2]$, horizontal flip, random contrast, and random brightness with a factor of 0.2 and random rotation with a degree uniformly drawn form $[-10^\circ,10^\circ]$. We adopt stochastic gradient descent~(SGD) to optimize the weights, with initial learning rate of 0.001 and momentum of 0.9.  Cosine scheduling with a period of  10 is used to  adjust  the  learning  rate  across  the  training  process. 

\paragraph{Results}

\begin{table}[htbp]
    \centering
        \caption{Performance of DenseNet-169 on the test set.}
    \begin{tabular}{lccc}
    \toprule
    Train data & Accuracy(\%) & F1-score(\%) & AUC(\%)
        \\ \midrule 
        COVID-CT-349 & 79.5 & 76.0 & 90.1\\
        COVID-CT-349 with lung mask only& 85.0  &85.9 & 92.8\\ \midrule
       COVID-Seg & 69.8 & 61.1 & 86.9 \\
       COVID-Seg with lung mask only &  68.0 & 74.8 & 85.8\\ 
        COVID-Seg with lesion mask only & 72.4& 77.5& 95.0    
%         COVID-Seg with lung and lesion masks & 69.8  & 76.5 & 90.9
        \\  \midrule
        COVID-CT-118 & 57.8& 36.3 & 75.2\\
        COVID-CT-118 with lung mask only & 73.3 & 66.4 & 86.2\\\midrule
        Combination & 74.5 & 70.1 & 89.8\\
        Combination with lung mask only& 86.2 & 87.2 & 97.6\\  
         Combination with lesion mask only & 83.3 & 84.6 & 94.8 \\
         Combination with lung and lesion masks & 87.1 & 88.1 & 95.2\\
         \bottomrule
    \end{tabular}
    \label{tab:mask-in}
\end{table}

Table~\ref{tab:mask-in} shows the performance of DenseNet-169 on test data. ``Combination" denotes the combined training dataset of COVID-CT-349 and COVID-Seg. DenseNet-169 is pretrained on ImageNet via transfer learning, without finetuning using CSSL. From this table, we make the following observations. First, incorporating lung masks can greatly improve performance. This is evident from the comparison of (1) ``COVID-CT-349 with lung mask only" and ``COVID-CT-349",  (2) ``COVID-CT-118 with lung mask only" and ``COVID-CT-118", and (3) ``Combination with lung mask only" and ``Combination". On COVID-Seg, incorporating  lung masks substantially improves F1, though the accuracy and AUC are sacrificed a bit. Second, incorporating lesion masks can greatly improve performance. This is evident from the comparison of (1) ``COVID-Seg with lesion mask only" and ``COVID-Seg" and (2) ``Combination with lesion mask only" and ``Combination". Third, on the combined dataset, using both lung masks and lesion masks achieve better performance than using lung masks only or using lesion masks only. Overall, these results demonstrate the usefulness of lung masks and lesions masks in improving diagnosis performance.

\begin{table}[h]
    \centering
    %\vspace{-0.6cm}
        \caption{Comparison of different pretraining methods.}
    \begin{tabular}{lccc}
    \toprule
    Initialization  & Accuracy(\%) & F1-score(\%) & AUC(\%)
        \\ \midrule 
        Random initialization & 83.0 & 83.2 & 94.1
        \\ 
        Transfer learning (TL) & 87.1 & 88.1 & 95.2
        \\  
        TL+CSSL & 89.1 & 89.6 & 98.1
        \\ \bottomrule 
    \end{tabular}
    \label{tab:pretraining}
\end{table}

Table~\ref{tab:pretraining} shows the testing performance of DenseNet-169 trained on the combined dataset, under different pretraining methods. From this table, we make the following observations. First, CSSL+TL achieves better performance than transfer learning (TL). This demonstrates the effectiveness of CSSL in improving the learned representations. Second, TL performs better than random initialization. By incorporating lung masks and lesion mask via multi-task learning and leveraging CSSL pretraining, our method achieves an F1 of 0.90, an AUC of 0.98, and an accuracy of 0.89. The senior radiologist has confirmed that models with such performance are clinically useful.

%Table~\ref{tab:5compare} shows the results of these five methods. From this table, we make the following observations. First, comparing JCS and DenseNet-169 on Combination, we can see that performing joint classification and segmentation can greatly improve performance. JCS on Combination achieves an accuracy of 83.3\%, an F1 score of 84.6, and an AUC of 94.8\%. This is because JCS incorporates additional supervision, which is the labeled regions of lesions, for model training. Second, JCS on Combination achieves substantially better accuracy and F1  than JCS on COVID-Seg, which further demonstrates the usefulness of our COVID-CT dataset. Third, JCS on COVID-Seg works better than DenseNet-169 on COVID-Seg. This further shows the effectiveness of performing segmentation and classification jointly. 

%Table~\ref{} shows the performance of models trained using lung masks and without using lung masks, under different settings of training data. As can be seen, for each setting of training data, using lung masks can greatly improve performance. This demonstrates the usefulness of lung masks. Without using lung masks, the performance on Combination is worse than that on COVID-CT-349. By using lung masks, the performance on Combination outperforms that on COVID-CT-349. 

%Table~\ref{} shows the performance of using both lung masks and lesion masks. As can be seen, it is better than using lung mask only or using lesion mask only. 

\section{Related works}
\subsection{Deep learning based diagnosis of COVID-19}
Since the outbreak of COVID-19, there have been increasing efforts on developing deep learning methods to perform screening of COVID-19 based on medical images such as CT scans and chest X-rays. Wu et al. established an early-screening model based on multiple CNN models to classify CT scans of patients with COVID-19~\citep{xu2020deep}. Wang et al. proposed a 3D deep CNN to detect COVID-19~\citep{zheng2020deep} using chest CT slices. Chowdhury et al. employed CNN to identify COVID-19 patients based on chest X-ray images~\citep{chowdhury2020can}. Several works have also applied 3D deep learning models to screen COVID-19 based on chest CT images~\citep{gozes2020rapid,li2020artificial}. Yang et al. developed a deep learning based CT diagnosis system (DeepPneumonia) to assist clinicians to identify patients with COVID-19~\citep{song2020deep}. Xu et al. developed a deep learning algorithm by modifying the inception transfer-learning model to provide clinical diagnosis ahead of the pathogenic test~\citep{wang2020deep}. Shi et al. employed the ``VB-Net” neural network to segment COVID-19 infection regions in CT scans~\citep{shan+2020lung}. Yu et al. constructed a system based on UNet++ for identification of COVID-19 from CT images~\citep{chen2020deep}. Shen et al. proposed an infection-size-aware Random Forest (iSARF) method  which can automatically categorize subjects into groups with different ranges of infected lesion sizes~\citep{shi2020large}. In most of these studies, the datasets are not publicly available.

\subsection{Datasets about COVID-19}

At present, few large-sized datasets with medical images on COVID-19 are publicly available due to privacy concerns and information blockade~\citep{cohen2020covid}. Existing datasets on COVID19 are mainly X-ray images~\citep{cohen2020covid,chowdhury2020can,COVID-19-3}. The Italian Society of Medical and Interventional Radiology (SIRM) provides chest X-rays and CT images of 68 Italian COVID-19 cases~\citep{COVID-19-1}. Moore et al. released a dataset of axial and coronal CTs from 59 COVID-19 cases at Radiopaedia~\citep{COVID-19-2}. Other data sources provide medical images of no more than 10 patients~\citep{Eurorad,Coronacases}. To deal with the lack of large-sized and open-source datasets containing CT images of COVID-19 cases, we built the COVID19-CT dataset.

\section{Conclusions}
We build a publicly available CT image dataset about COVID-19, to foster the development of AI methods for using CT to screen and test COVID-19 patients. The dataset contains 349 COVID-19 CT images from 216 patients and 463 non-COVID-19 CT images (used as negative training examples).  The utility of this dataset is confirmed by a senior radiologist who has intensively practiced diagnosis and treatment of COVID-19 patients. We also perform experimental studies to further verify the utility of this dataset. Using this dataset, we develop an  approach based on multi-task learning and contrastive self-supervised learning that achieves an F1 of 0.90, an AUC of 0.98, and an accuracy of 0.89, on a test set of original CT images donated by hospitals. The senior radiologist has confirmed that models with such performance are clinically useful.

\bibliographystyle{unsrt}
\bibliography{main}

%\appendix
%\section*{Appendix A.}

%Some more details about those methods, so we can actually reproducethem.  After the blind review period, you could link to a repository for the code also.  \emph{MLHC values both rigorous evaluation as well as reproduciblity.}

\end{document}